\documentclass[10.5pt,compsoc]{tst}
\usepackage{graphicx}
\usepackage{footmisc}
\usepackage{subfigure}
\usepackage{url}
\usepackage{multirow}
\usepackage[noadjust]{cite}
\usepackage{amsmath,amsthm}
\usepackage{amssymb,amsfonts}
\usepackage{booktabs}
\usepackage{color}
\usepackage{ccaption}
\usepackage{booktabs}
\usepackage{float}
\usepackage{fancyhdr}
\usepackage{caption}
\usepackage{xcolor,stfloats}
\usepackage{comment}
\graphicspath{{figures/}}
\usepackage{cuted}  
\usepackage{captionhack}
\usepackage{epstopdf}

\headevenname{\normalsize{\textbf{\emph{Tsinghua Science and Technology, xxxxx}} 20xx, 2x(x): xxx--xxx}}%
\headoddname{{\sf First author et al.:}\quad {\textbf{\emph{Click and Type Your Title Here ...}}}}%

\newtheoremstyle{mystyle}{0pt}{0pt}{\normalfont}{1em}{\bf}{}{1em}{}
\theoremstyle{mystyle}
\renewcommand\figurename{Fig.~}

\newtheorem{definition}{\textbf{Definition}}
\newtheorem{lemma}{\textbf{Lemma}}
\newtheorem{remark}{\textbf{Remark}}
\newtheorem{theorem}{\textbf{Theorem}}

\newcommand{\nop}[1]{}

\makeatletter
\renewcommand{\@biblabel}[1]{[#1]\hfill}
\makeatother

\begin{document}

\makeatletter
\newcommand\mysmall{\@setfontsize\mysmall{7}{9.5}}

\newenvironment{tablehere}
  {\def\@captype{table}}
  {}
\newenvironment{figurehere}
  {\def\@captype{figure}}
  {}

\thispagestyle{plain}%
\thispagestyle{empty}%

\let\temp\footnote
\renewcommand \footnote[1]{\temp{\normalsize #1}}
{}
\vspace*{-40pt}
\noindent{\normalsize\textbf{\scalebox{0.885}[1.0]{\makebox[5.9cm][s]
{TSINGHUA\, SCIENCE\, AND\, TECHNOLOGY}}}}

\vskip .2mm
{\normalsize
\textbf{
\hspace{-5mm}
\scalebox{1}[1.0]{\makebox[5.6cm][s]{%
I\hfill S\hfill S\hfill N\hfill{\color{white}%
l\hfill l\hfill}1\hfill0\hfill0\hfill7\hfill-\hfill0\hfill2\hfill1\hfill4
\hfill \color{white}{\quad 0\hfill ?\hfill /\hfill ?\hfill ?\quad p\hfill p\hfill  ?\hfill ?\hfill ?\hfill --\hfill ?\hfill ?\hfill ?}\hfill}}}}

\vskip .2mm
{\normalsize
\textbf{
\hspace{-5mm}
\scalebox{1}[1.0]{\makebox[5.6cm][s]{%
DOI:~\hfill~\hfill1\hfill0\hfill.\hfill2\hfill6\hfill5\hfill9\hfill9\hfill/\hfill T\hfill S\hfill T\hfill.\hfill2\hfill0\hfill x\hfill x\hfill.\hfill9\hfill0\hfill1\hfill0\hfill x\hfill x\hfill x}}}}

\vskip .2mm\noindent
{\normalsize\textbf{\scalebox{1}[1.0]{\makebox[5.6cm][s]{%
\color{black}{V\hfill o\hfill l\hfill u\hfill m\hfill%
e\hspace{0.356em}xx,\hspace{0.356em}N\hfill u\hfill%
m\hfill b\hfill e\hfill r\hspace{0.356em}x,\hspace{0.356em}%
x\hfill x\hfill x\hfill x\hfill x\hfill%
x\hfill x\hfill \hspace{0.356em}2\hfill0\hfill x\hfill x}}}}}\\

\begin{strip}
{\center
{\LARGE\textbf{A Predefined-Time Convergent and Noise-Tolerant Zeroing Neural Network Model for Time Variant Quadratic Programming With Application to Robot Motion Planning}}
\vskip 9mm}

{\center {\sf \large
Yi Yang, Xuchen Wang, Richard M. Voyles, and Xin Ma$^*$
}
\vskip 5mm}

\centering{
\begin{tabular}{p{160mm}}

{\normalsize
\linespread{1.6667} %
\noindent
\bf{Abstract:} {\sf
This paper develops a predefined-time convergent and noise-tolerant fractional-order zeroing neural network (PTC-NT-FOZNN) model, innovatively engineered to tackle time-variant quadratic programming (TVQP) challenges. The PTC-NT-FOZNN, stemming from a novel iteration within the variable-gain ZNN spectrum, known as FOZNNs, features diminishing gains over time and marries noise resistance with predefined-time convergence, making it ideal for energy-efficient robotic motion planning tasks. The PTC-NT-FOZNN enhances traditional ZNN models by incorporating a newly developed activation function that promotes optimal convergence irrespective of the model’s order. When evaluated against six established ZNNs, the PTC-NT-FOZNN, with parameters $0<\alpha\leq 1$, demonstrates enhanced positional precision and resilience to additive noises, making it exceptionally suitable for TVQP tasks. Thorough practical assessments, including simulations and experiments using a Flexiv Rizon robotic arm, confirm the PTC-NT-FOZNN's capabilities in achieving precise tracking and high computational efficiency, thereby proving its effectiveness for robust kinematic control applications.}
}

\end{tabular}
}
\vskip 6mm

\vskip -3mm
\small\end{strip}

\thispagestyle{plain}%
\thispagestyle{empty}%
\makeatother
\pagestyle{tstheadings}

\begin{figure}[b]
\vskip -6mm
\begin{tabular}{p{44mm}}
\toprule\\
\end{tabular}
\vskip -4.5mm
\noindent
\setlength{\tabcolsep}{1pt}
\begin{tabular}{p{1.5mm}p{79.5mm}}
$\bullet$& Yi Yang, Xuchen Wang and Xin Ma are with the the Multi-Scale Medical Robotics Center and the Department of Mechanical and Automation Engineering, The Chinese University of Hong Kong, Hong Kong 999077, China. E-mail: yiyang@mrc-cuhk.com; xcwang@mae.cuhk.edu.hk; xinma001@cuhk.edu.hk\\
$\bullet$& Richard M. Voyles is with the School of Engineering Technology, Purdue University, West Lafayette 47907, IN, USA. E-mail:
rvoyles@purdue.edu \\
$\sf{*}$&
To whom correspondence should be addressed. \\
          &          Manuscript received: 2024-06-26;
          accepted: 2024-10-17

\end{tabular}
\end{figure}\large

\section{Introduction}
\label{s:introduction}
\noindent
Quadratic programming (QP) is a fundamental aspect of various fields in science and engineering, notably within robotic kinematic control and dynamics modeling \cite{1,2,3}. These problems often take the form of time-variant quadratic programming (TVQP) \cite{4,5,6}, prompting the development of efficient algorithms \cite{7,8,9} for their resolution. Among these, recurrent neural networks (RNNs), celebrated for their adeptness at handling time-series data, have emerged as key tools for addressing QPs, especially because they are compatible with parallel computing environments \cite{10,11,12}. Nonetheless, gradient neural networks (GNNs), a type of RNN, are generally better suited for static problems and have faced challenges such as delayed error corrections when applied to TVQP tasks \cite{10,13,14}.

To overcome these shortcomings, the zeroing neural network (ZNN) models have been introduced. These models are designed to address TVQP challenges effectively \cite{15,16} and have demonstrated rapid convergence rates in the kinematic control of robotics, greatly influenced by the selection of activation functions \cite{17,18}. Recent advancements have led to the development of ZNN models that achieve finite-time, fixed-time, and predefined-time convergence \cite{19,20,21}. However, the application of high constant gains or increasing variable gains, which can enhance the convergence rate and accuracy of ZNNs, presents substantial difficulties in practical settings due to power limitations and other operational constraints \cite{22,23}. To illustrate, in \cite{24}, a neural network running on a mobile device for real-time tasks shows a direct correlation between gain increase and power demand, leading to significant battery drain. Further research \cite{25} examines how gain adjustments in recurrent neural networks aim to enhance energy efficiency. This work emphasizes the adaptive changes in RNN operation to minimize energy use concerning action potentials and synaptic activity. Studies \cite{26} and \cite{27} further validate the substantial effects of gain settings on the deployment of spiking neural networks within energy-sensitive environments. Moreover, ZNN models are often sensitive to noise, including measurement errors and computational inaccuracies in hardware systems, which can adversely affect their performance and reliability in real-world applications \cite{21,28}, thereby underscoring a significant limitation in their practical deployment under time-varying gain conditions.

Leveraging the unique properties of conformable fractional derivatives \cite{29}, which endows a time-diminishing variable gain to the standard ZNN model, this paper proposes a predefined-time convergent and noise-tolerant fractional-order zeroing neural network (PTC-NT-FOZNN) model. This model is designed to solve TVQP problems with equality and inequality constraints. The PTC-NT-FOZNN is distinguished by a new activation function that ensures noise immunity and predefined-time convergence, regardless of the model's order. Numerical evaluation demonstrates that the PTC-NT-FOZNN outperforms six established ZNNs in terms of convergence efficiency. Additionally, its effectiveness as an inverse-kinematics solver for robotic motion planning has been rigorously verified. The contributions of this paper include:

(1) The introduction of a novel predefined-time convergent and anti-noise activation function for a variable-gain ZNN architecture with hardware power  efficiency, termed FOZNNs, marks a significant step forward in addressing TVQP challenges.

(2) A rigorous proof establishing the PTC-NT-FOZNN’s capabilities for predefined-time convergence and noise tolerance, underscoring its suitability for inverse kinematic control of robotic arms.

(3) The model's superior convergence accuracy and robustness against additive noises, as compared to six other ZNN models, demonstrated through both theoretical analysis and practical tests using a Flexiv Rizon robotic arm.

The structure of this paper is as follows: Section 2 delves into TVQP problems and the foundational theories crucial for developing traditional ZNNs. Section 3 elaborates on the PTC-NT-FOZNN model, including a thorough analysis of its convergence properties. Section 4 provides numerical evidence demonstrating the model's effectiveness in solving TVQPs and explores its implementation in robotic kinematic control through both simulations and real-world experiments. Finally, Section 5 summarizes the findings and conclusions of the study.

\renewcommand{\arraystretch}{1.2}
\begin{table}[H]
\caption{Explanation for mathematical notations.}\label{tbl1}
\begin{tabular}{ll}
\\
\hline Notation & Explanation \\
\hline$\|\boldsymbol{x}\|$ & 2-norm of a vector $x \in \mathbb{R}^n$ \\
$\boldsymbol{x}_1 \circ \boldsymbol{x}_2$ & Hadamard product of two vectors $x_1$ and $x_2$ \\
$\boldsymbol{x}_1 \oslash \boldsymbol{x}_2$ & Hadamard division of two vectors $x_1$ and $x_2$ \\
$C_\alpha(\boldsymbol{\epsilon})(t)$ & Conformable fractional derivative of $\boldsymbol{\epsilon}(t)$ \\
$\boldsymbol{\epsilon}(t)$ & A vector function representing residual error \\
$\dot{\boldsymbol{\epsilon}}(t)$ & First derivative for the residual error \\
\hline
\end{tabular}
\end{table}

\section{Preliminary Theory}
\label{s:Experimental}
\noindent
This section outlines the formulation of the TVQP problem, foundational preliminaries including relevant definitions and lemmas, alongside the traditional ZNN model and its variation under noise influence.
\subsection{Problem Formulation}
\noindent
A recurrent neural network (RNN) model is defined in \cite{30}, as follows:
\begin{equation}
\boldsymbol{y}(t)=\phi(t, \boldsymbol{y}(t)), t \in[0, \infty)
\label{eq1}
\end{equation}
where $\boldsymbol{y}(t) \in \mathbb{R}^{n}$ is the state vector of the RNN, and $\phi(\cdot)$ denotes the activation function for the RNN.

This study focuses on devising a fractional-order RNN model with favorable convergence and anti-noise characteristics to effectively solve the following TVQP problems involving equality and inequality constraints,
\begin{equation}
\begin{array}{ll}
\min & \boldsymbol{y}^{\mathrm{T}}(t) \Omega(t) \boldsymbol{y}(t) / 2+\boldsymbol{p}^{\mathrm{T}}(t) \boldsymbol{y}(t) \\
\text { s.t. } & A(t) \boldsymbol{y}(t)=\boldsymbol{b}(t) \\
& C(t) \boldsymbol{y}(t) \leq \boldsymbol{d}(t)
\end{array}
\label{eq2}
\end{equation}
where $\Omega(t) \in \mathbb{R}^{n \times n}$ denotes a positive semi-definite matrix; \(A(t) \in \mathbb{R}^{m \times n}\) and \(C(t) \in \mathbb{R}^{l \times n}\) represent two matrices with full row rank; \(\boldsymbol{p}(t) \in \mathbb{R}^{n}, \boldsymbol{b}(t) \in\) \(\mathbb{R}^{m}\) and \(\boldsymbol{d}(t) \in \mathbb{R}^{l}\) are associated vectors.

If a solution to the TVQP problem (\ref{eq2}) exists, it is termed as the Karush-Kuhn-Tucker (KKT) point.
\begin{lemma}
\cite{31} For the TVQP problem (\ref{eq2}), \(\boldsymbol{y}(t) \in\) \(\mathbb{R}^{n}\) is the KKT point if, for any \(\varepsilon \rightarrow 0_{+}\)there exist Lagrangian multipliers \(\boldsymbol{\lambda}_{\mathbf{1}}(t) \in \mathbb{R}^{m}\) and \(\boldsymbol{\lambda}_{\mathbf{2}}(t) \in \mathbb{R}^{l}\) to make the following conditions hold:
\begin{equation}
\left\{\begin{array}{l}
\Omega(t) \boldsymbol{y}(t)+\boldsymbol{p}(t)+A^{\mathrm{T}}(t) \boldsymbol{\lambda}_{\mathbf{1}}(t)+C^{\mathrm{T}}(t) \boldsymbol{\lambda}_{\mathbf{2}}(t)=0 \\
A(t) \boldsymbol{y}(t)-\boldsymbol{b}(t)=0 \\
\phi_{F B}^{\varepsilon}\left(\boldsymbol{d}(t)-C(t) \boldsymbol{y}(t), \boldsymbol{\lambda}_{\mathbf{2}}(t)\right)=0
\end{array}\right.
\label{eq3}
\end{equation}
where \(\phi_{F B}^{\varepsilon}\) denotes the perturbed Fischer-Burmeister (FB) function \cite{31}, which is defined as
\begin{equation}
\phi_{F B}^{\varepsilon}(\boldsymbol{a}, \boldsymbol{b})=\boldsymbol{a}+\boldsymbol{b}-\sqrt{\boldsymbol{a} \circ \boldsymbol{a}+\boldsymbol{b} \circ \boldsymbol{b}+\varepsilon}, \varepsilon \rightarrow 0_{+}
\label{eq4}
\end{equation}
where \(\boldsymbol{a}, \boldsymbol{b}\) are vectors of identical dimensions, \(\varepsilon\) represents a minor perturbation, the symbol \(\circ\) denotes the Hadamard product, and the square root operation is applied element-wise to the vector.
\label{lem1}
\end{lemma}

Aligned with the locally Lipschitz continuity condition outlined in \cite{32}, this research assumes the uniqueness of the optimal solution for the TVQP (\ref{eq2}). As substantiated by Lemma 1, (\ref{eq2}) can be effectively solved by finding the solution to the following timevariant linear matrix equality (TVLME),
\begin{equation}
\boldsymbol{\epsilon}(t)=E(t) \boldsymbol{z}(t)+\boldsymbol{f}(t)=0
\label{eq5}
\end{equation}
where \(\boldsymbol{z}(t)=\left[\boldsymbol{y}^{\mathrm{T}}(t), \boldsymbol{\lambda}_{\mathbf{1}}^{\mathrm{T}}(t), \boldsymbol{\lambda}_{\mathbf{2}}^{\mathrm{T}}(t)\right]^{\mathrm{T}} \in \mathbb{R}^{n+m+l}\), $\boldsymbol{f}(t)=\left[
\boldsymbol{p}^{\mathrm{T}}(t),
-\boldsymbol{b}^{\mathrm{T}}(t),
\boldsymbol{d}^{\mathrm{T}}(t)-\boldsymbol{m}^{\mathrm{T}}(t)
\right]^{\mathrm{T}}$, and
\begin{equation*}
E(t)=\left[\begin{array}{ccc}
\Omega(t) & A^{\mathrm{T}}(t) & C^{\mathrm{T}}(t) \\
A(t) & 0 & 0 \\
-C(t) & 0 & I
\end{array}\right], 
\end{equation*}
with \(\boldsymbol{\epsilon}(t)\) denoting the residual error function. In addition, \(\boldsymbol{m}(t)=\sqrt{\boldsymbol{n}(t) \circ \boldsymbol{n}(t)+\boldsymbol{\lambda}_{\mathbf{2}}(t) \circ \boldsymbol{\lambda}_{\mathbf{2}}(t)+\varepsilon}\) and \(\boldsymbol{n}(t)=\boldsymbol{d}(t)-C(t) \boldsymbol{y}(t)\). The resolution of the TVQP problem (\ref{eq2}) is converted into the addressing of the TVLME (\ref{eq5}).
\subsection{Traditional ZNN model}
\noindent
The zeroing neural network (ZNN) model is formulated as
\begin{equation}
\dot{\boldsymbol{\epsilon}}(t)=-\gamma \Pi(\boldsymbol{\epsilon}(t))
\label{eq6}
\end{equation}
where \(\gamma>0\) represents the constant gain factor, and \(\Pi(\cdot): \mathbb{R}^{n+m+l} \rightarrow \mathbb{R}^{n+m+l}\) is the activation function.

The evolution dynamics (\ref{eq6}) can be expanded as
\begin{equation}
P(t) \dot{\boldsymbol{z}}(t)=-Q(t) \boldsymbol{z}(t)-\boldsymbol{\rho}(t)-\gamma \Pi(\boldsymbol{\epsilon}(t))
\label{eq7}
\end{equation}
where
\begin{equation*}\begin{aligned}
P(t) & =\left[\begin{array}{ccc}
\Omega(t) & A^{\mathrm{T}}(t) & C^{\mathrm{T}}(t) \\
A(t) & 0 & 0 \\
\left(\Lambda_{1}(t)-I\right) C(t) & 0 & I-\Lambda_{2}(t)
\end{array}\right] \\
Q(t) & =\left[\begin{array}{ccc}
\dot{\Omega}(t) & \dot{A}^{\mathrm{T}}(t) & \dot{C}^{\mathrm{T}}(t) \\
\dot{A}(t) & 0 & 0 \\
\left(\Lambda_{1}(t)-I\right) \dot{C}(t) & 0 & 0
\end{array}\right] \\
\boldsymbol{\rho}(t) & =\left[\begin{array}{c}
\dot{\boldsymbol{p}}(t) \\
-\dot{\boldsymbol{b}}(t) \\
\left(I-\Lambda_{1}(t)\right) \dot{\boldsymbol{d}}(t)
\end{array}\right]
\end{aligned}
\end{equation*}
with \(\Lambda_{1}(t)=\operatorname{diag}(\boldsymbol{n}(t) \oslash \boldsymbol{m}(t))\) and \(\Lambda_{2}(t)=\) \(\operatorname{diag}\left(\boldsymbol{\lambda}_{\mathbf{2}}(t) \oslash \boldsymbol{m}(t)\right)\), where \(\oslash\) signifies the Hadamard division and \(\operatorname{diag}(\cdot)\) denotes a diagonal matrix.

In practical ZNN deployments, extraneous noises are unavoidable due to computational roundoff errors, measurement, and differentiation-induced inaccuracy. A noise-polluted ZNN configured to address the TVQP problem (2) is described as follows:
\begin{equation}
P(t) \dot{\boldsymbol{z}}(t)=-Q(t) \boldsymbol{z}(t)-\boldsymbol{\rho}(t)-\gamma \Pi(\boldsymbol{\epsilon}(t))+\boldsymbol{\delta}(t)
\label{eq8}
\end{equation}
where \(\boldsymbol{\delta}(t)\) signifies the additive noise term.

Before discussing the design and convergence analysis of the fractional-order neural model, we first introduce some preliminary definitions and lemmas.
\begin{definition}
    \cite{33,34,35} For an RNN model (\ref{eq1}) with \(\boldsymbol{y}_{0}=\boldsymbol{y}(0)\) representing its initial state, if \(\boldsymbol{y}(t)=0\) is assumed as the equilibrium point of the system, then
    
    (1) Local finite-time stability occurs if there exists an open set \(\Gamma\) around the origin and a locally bounded settling-time function \(T: \Gamma \backslash\{0\} \rightarrow \mathbb{R}_{+} \cup\{0\}\), such that for any trajectory \(\boldsymbol{y}\left(t, \boldsymbol{y}_{0}\right)\) originating from \(\boldsymbol{y}_{0} \in\Gamma \backslash\{0\}\), convergence to the origin is guaranteed for all \(t \geq T\left(\boldsymbol{y}_{0}\right)\).
    
    (2) Local fixed-time stability occurs when it is locally finite-time stable with an additional condition: there exists a constant \(T_{\max }>0\) ensuring that \(T\left(\boldsymbol{y}_{0}\right) \leq\) \(T_{\max }\) for all \(\boldsymbol{y}_{0} \in \Gamma\). Global fixed-time stability is achieved if \(\Gamma=\mathbb{R}^{n}\).
    
    (3) Local predefined-time stability is defined by its local fixed-time stability coupled with the condition that \(T\left(\boldsymbol{y}_{0}\right) \leq t_{c}\) for all \(\boldsymbol{y}_{0} \in \Gamma\), where \(t_{c}>0\) is the predefined time constant. Global predefined-time stability is affirmed if \(\Gamma=\mathbb{R}^{n}\).
\end{definition}
\begin{definition}
    \cite{29,36} The conformable fractional derivative of a function \(g(t)\), which is differentiable up to the \(n\)-th order, is defined as
    \begin{equation}
    C_{\alpha}(g)(t)=\lim _{\omega \rightarrow 0} \frac{g^{(n)}\left(t+\omega t^{n+1-\alpha}\right)-g^{(n)}(t)}{\omega}
    \label{eq9}
    \end{equation}
where \(\alpha\in (n, n+1]\) is the order of the conformable fractional derivative operator, with \(n \geq 0\) being an integer. A function \(g(t)\) is deemed \(\alpha\)-differentiable at a point \(t>0\) if the operator \(C_{\alpha}(g)(t)\) is defined at that point.
\end{definition}
\begin{lemma}
    \cite{29} For \(0<\alpha \leq 1\), and given \(u(t)\) and \(v(t)\) are two \(\alpha\)-differentiable functions at a point \(t>0\), the following properties hold:\\
(1) \(C_{\alpha}(a u+b v)=a C_{\alpha}(u)+b C_{\alpha}(v)\), for \(a, b \in \mathbb{R}\).\\
(2) \(C_{\alpha}\left(t^{a}\right)=a t^{a-\alpha}\) for any \(a \in \mathbb{R}\).\\
(3) \(C_{\alpha}(C)=0\) for any constant function \(u(t)=C\).\\
(4) \(C_{\alpha}(u v)=u C_{\alpha}(v)+v C_{\alpha}(u)\).\\
(5) \(C_{\alpha}(u / v)=\left(v C_{\alpha}(u)-u C_{\alpha}(v)\right) / v^{2}\).\\
(6) \(C_{\alpha}(u)(t)=t^{1-\alpha} u^{\prime}(t)\) for any function \(u(t)\).
\end{lemma}

\section{Scheme Design and Main Results}
\label{s:section3}
\noindent
This section details the design of the predefined-time convergent and noise-tolerant fractional-order ZNN (PTC-NT-FOZNN) model, offering theoretical justification into its predefined-time convergence and noise resistance characteristics.
\subsection{Design of PTC-NT-FOZNN Model}
\noindent
A fractional-order ZNN (FOZNN) model \cite{37} is derived by replacing \(\dot{\boldsymbol{\epsilon}}(t)\) in (\ref{eq6}) with the conformable fractional derivative of \(\boldsymbol{\epsilon}(t)\),
\begin{equation}
    C_{\alpha}(\boldsymbol{\epsilon})(t)=t^{1-\alpha} \dot{\boldsymbol{\epsilon}}(t)=-\gamma \Pi(\boldsymbol{\epsilon}(t))
    \label{eq10}
\end{equation}
where the order \(\alpha\) is prescribed to be \(0<\alpha \leq 1\), and the term \(t^{1-\alpha}\) is derived from the property (6) in Lemma 2 for the conformable fractional derivative. Correspondingly, the noise-polluted FOZNN model is given as
\begin{equation}
    \dot{\boldsymbol{\epsilon}}(t)=-\gamma t^{\alpha-1} \Pi(\boldsymbol{\epsilon}(t))+\boldsymbol{\delta}(t)
    \label{eq11}
\end{equation}

To imbue the FOZNN model (\ref{eq10}) and (\ref{eq11}) with predefined-time convergence capabilities, we propose a predefined-time stabilizer as the activation function,
\begin{equation}
    \Pi(\boldsymbol{\epsilon})=\frac{\pi}{2 \kappa \gamma t_{c}^{\alpha}}\left(\|\boldsymbol{\epsilon}\|^{1-\kappa}+\|\boldsymbol{\epsilon}\|^{1+\kappa}\right) \frac{\boldsymbol{\epsilon}}{\|\boldsymbol{\epsilon}\|}+\frac{\zeta}{\gamma t^{\alpha-1}} \frac{\boldsymbol{\epsilon}}{\|\boldsymbol{\epsilon}\|}
    \label{eq12}
\end{equation}
where \(\kappa\) is a positive constant with \(0<\kappa<1\), \(t_{c}\) represents the predefined time, and \(\zeta\) is an arbitrary upper bound for the additive noise.
\subsection{Main Results in Convergence Analysis}
\noindent
This section elaborates on the convergence and robustness of the PTC-NT-FOZNN, activated by (\ref{eq12}). Theoretical analyses are provided to assure predefined-time convergence for FOZNNs (\ref{eq10}) and (\ref{eq11}), under activation function (\ref{eq12}).
\begin{theorem}
    For the TVQP problem (\ref{eq2}) or, equivalently, the TVLME (\ref{eq5}), using the activation function (\ref{eq12}) ensures that the neural state \(\boldsymbol{z}(t)\) of the FOZNN (\ref{eq10}) without additive noise, starting from an initial state \(z_{0}\) near the theoretical initial state \(\boldsymbol{z}_{0}^{*}\), will converge to the theoretical solution \(\boldsymbol{z}^{*}(t)\) in the predefined time \(t_{c}\).
\end{theorem}
\begin{IEEEproof}
    For \(t<t_{c}\), substituting the activation function (\ref{eq12}) into the FOZNN model (\ref{eq10}) produces:
\begin{equation}
    \begin{aligned}
& \dot{\boldsymbol{\epsilon}}(t)=-\frac{\pi t^{\alpha-1}}{2 \kappa t_{c}^{\alpha}}\left(\|\boldsymbol{\epsilon}(t)\|^{1-\kappa}+\|\boldsymbol{\epsilon}(t)\|^{1+\kappa}\right) \frac{\boldsymbol{\epsilon}(t)}{\|\boldsymbol{\epsilon}(t)\|} \\
&-\zeta \frac{\boldsymbol{\epsilon}(t)}{\|\boldsymbol{\epsilon}(t)\|}
\end{aligned}
\label{eq13}
\end{equation}

Consider a Lyapunov functional \(V(t)=\|\boldsymbol{\epsilon}(t)\|\), then one obtains:
\begin{equation}
    \begin{aligned}
& \dot{V}(t)=\frac{\partial V}{\partial \boldsymbol{\epsilon}} \dot{\boldsymbol{\epsilon}}(t)=\frac{\boldsymbol{\epsilon}^{\mathrm{T}}(t)}{\|\boldsymbol{\epsilon}(t)\|} \dot{\boldsymbol{\epsilon}}(t) \\
& =-\frac{\pi}{2 \kappa t_c}\left(\frac{t}{t_c}\right)^{\alpha-1}\left(\|\boldsymbol{\epsilon}(t)\|^{1-\kappa}+\|\boldsymbol{\epsilon}(t)\|^{1+\kappa}\right)-\zeta \\
& \leq-\frac{\pi}{2 \kappa t_c}\left(V^{1-\kappa}+V^{1+\kappa}\right)<0
\end{aligned}
\label{eq14}
\end{equation}

According to the Lyapunov stability theorem, the origin is globally finite-time stable. Integrating both sides of the differential inequality (\ref{eq14}) provides an upper bound for the settling-time function:
\begin{equation}
   \begin{aligned}
T\left(\epsilon_0\right) & \leq-\frac{2 \kappa t_c}{\pi} \int_{V\left(\epsilon_0\right)}^0 \frac{\mathrm{~d} V}{V^{1-\kappa}+V^{1+\kappa}} \\
& \leq \frac{2 t_c}{\pi} \arctan \left(V\left(\epsilon_0\right)^\kappa\right) \leq t_c
\end{aligned}
\label{eq15}
\end{equation}

For \(t \geq t_{c}\), the derivative of Lyapunov functional yields:
\begin{equation}
\begin{gathered}
\dot{V}(t)=-\frac{\pi}{2 \kappa t_c}\left(\frac{t}{t_c}\right)^{\alpha-1}\left(\|\boldsymbol{\epsilon}(t)\|^{1-\kappa}+\|\boldsymbol{\epsilon}(t)\|^{1+\kappa}\right) \\
-\zeta<0
\end{gathered}
\label{eq16}
\end{equation}
which is permanently negative, ensuring that \(\boldsymbol{\epsilon}(t)=0\) is always upheld for all \(t>t_{c}\). This suggests the predefined-time convergence of the FOZNN (\ref{eq10}).
\end{IEEEproof}

\renewcommand{\arraystretch}{1.5}
\begin{table*}
\caption{Comparison among some representative ZNN models.}\label{tbl2}
\begin{tabular}{lllll}
\\
\hline
Schemes & Gain & Activation function & Convergence & Noise rejection \\
\hline
Scheme in \cite{11} & Constant \(\gamma\) & \(\Pi(\boldsymbol{\epsilon})=\exp \left(|\boldsymbol{\epsilon}|^{r}\right)|\boldsymbol{\epsilon}|^{1-r} \operatorname{sign}(\boldsymbol{\epsilon}) / r\) & Predefined-time & No \\
Scheme in \cite{18} & Constant \(\gamma\) & \( \Pi(\boldsymbol{\epsilon})= \begin{cases}\frac{\exp (\boldsymbol{\epsilon})-1}{\left(t_{c}-t\right) \exp (\boldsymbol{\epsilon})}, & t<t_{c} \\ \boldsymbol{\epsilon}, & t \geq t_{c}\end{cases} \) & Predefined-time & No \\
Scheme in \cite{19} & Constant \(\gamma\) & \(\Pi(\boldsymbol{\epsilon})=\boldsymbol{\epsilon}+|\boldsymbol{\epsilon}|^{r} \operatorname{sign}(\boldsymbol{\epsilon})+|\boldsymbol{\epsilon}|^{1 / r} \operatorname{sign}(\boldsymbol{\epsilon})\) & Fixed-time & No \\
Scheme in \cite{20} & Constant \(\gamma\) & \(\Pi(\boldsymbol{\epsilon})=\exp \left(|\boldsymbol{\epsilon}|^{r}\right)|\boldsymbol{\epsilon}|^{1-r} \operatorname{sign}(\boldsymbol{\epsilon}) / r+\zeta \operatorname{sign}(\boldsymbol{\epsilon})\) & Predefined-time & Yes \\
Scheme in \cite{38} & Constant \(\gamma\) & \(\Pi(\boldsymbol{\epsilon})=\boldsymbol{\epsilon}+|\boldsymbol{\epsilon}|^{r} \operatorname{sign}(\boldsymbol{\epsilon})\) & Finite-time & No \\
Scheme in \cite{39} & Constant \(\gamma\) & \( \Pi(\boldsymbol{\epsilon})=\left\{\begin{array}{lr} \frac{\boldsymbol{\epsilon}}{t_{c}-t}, & t<t_{c} \\ \boldsymbol{\epsilon}+|\boldsymbol{\epsilon}|^{r} \operatorname{sign}(\boldsymbol{\epsilon})+\xi \operatorname{sign}(\boldsymbol{\epsilon}), & t \geq t_{c} \end{array}\right. \) & Predefined-time & Yes \\
PTC-NT-FOZNN & \(\gamma(t)=\gamma t^{\alpha-1}\) & \( \Pi(\boldsymbol{\epsilon})=\frac{\pi}{2 \kappa \gamma t_{c}^{\alpha}}\left(\|\boldsymbol{\epsilon}\|^{1-\kappa}+\|\boldsymbol{\epsilon}\|^{1+\kappa}\right) \frac{\boldsymbol{\epsilon}}{\|\boldsymbol{\epsilon}\|}+\frac{\zeta}{\gamma t^{\alpha-1}} \frac{\boldsymbol{\epsilon}}{\|\boldsymbol{\epsilon}\|} \) & Predefined-time & Yes \\
\hline
\end{tabular}
\end{table*}

\begin{theorem}
    For the TVQP problem (\ref{eq2}) or, equivalently, the TVLME (\ref{eq5}), using the activation function (\ref{eq12}) ensures that the neural state \(\boldsymbol{z}(t)\) of the FOZNN (\ref{eq11}) with additive noise, starting from an initial state \(z_{0}\) near the theoretical initial state \(z_{0}^{*}\), will converge to the theoretical solution \(\boldsymbol{z}^{*}(t)\) in the predefined time \(t_{c}\).
\end{theorem}
\begin{IEEEproof}
    For \(t<t_{c}\), inserting the activation function (\ref{eq12}) into the FOZNN model (\ref{eq11}) yields:
\begin{equation}
    \begin{gathered}
\dot{\boldsymbol{\epsilon}}(t)=-\frac{\pi t^{\alpha-1}}{2 \kappa t_c^\alpha}\left(\|\boldsymbol{\epsilon}(t)\|^{1-\kappa}+\|\boldsymbol{\epsilon}(t)\|^{1+\kappa}\right) \frac{\boldsymbol{\epsilon}(t)}{\|\boldsymbol{\epsilon}(t)\|} \\
-\zeta \frac{\boldsymbol{\epsilon}(t)}{\|\boldsymbol{\epsilon}(t)\|}+\boldsymbol{\delta}(t)
\end{gathered}
\label{eq17}
\end{equation}

Consider a Lyapunov functional \(V(t)=\|\boldsymbol{\epsilon}(t)\|\), then one has
\begin{equation}
    \begin{aligned}
& \dot{V}(t)=-\frac{\pi}{2 \kappa t_{c}}\left(\frac{t}{t_{c}}\right)^{\alpha-1}\left(\|\boldsymbol{\epsilon}(t)\|^{1-\kappa}+\|\boldsymbol{\epsilon}(t)\|^{1+\kappa}\right) \\
&-\zeta+\frac{\boldsymbol{\epsilon}^{\mathrm{T}}(t) \boldsymbol{\delta}(t)}{\|\boldsymbol{\epsilon}(t)\|} \\
& \leq-\frac{\pi}{2 \kappa t_{c}}\left(V^{1-\kappa}+V^{1+\kappa}\right)-(\zeta-\|\boldsymbol{\delta}(t)\|)
\end{aligned}
\label{eq18}
\end{equation}

Since \(\zeta>\|\boldsymbol{\delta}(t)\|\) strictly holds, integrating on both sides of (\ref{eq18}) results in \(T\left(\epsilon_{0}\right) \leq t_{c}\).

For \(t \geq t_{c}\), one has
\begin{equation}
\begin{gathered}
\dot{V}(t)=-\frac{\pi}{2 \kappa t_{c}}\left(\frac{t}{t_{c}}\right)^{\alpha-1}\left(\|\boldsymbol{\epsilon}(t)\|^{1-\kappa}+\|\boldsymbol{\epsilon}(t)\|^{1+\kappa}\right) \\
-\zeta+\frac{\boldsymbol{\epsilon}^{\mathrm{T}}(t) \boldsymbol{\delta}(t)}{\|\boldsymbol{\epsilon}(t)\|}<0
\end{gathered}
\label{eq19}
\end{equation}

Then, the negativeness of \(\dot{V}(t)\) indicates that the FOZNN (\ref{eq11}) under the activation function (\ref{eq12}) is predefined-time convergent.
\end{IEEEproof} 
\begin{remark}
    From the proofs of the two theorems, it is evident that in steady state, the residual error is bounded as follows:
    \begin{equation}
        \|\boldsymbol{\epsilon}(\infty)\|=V\left(\epsilon_{0}\right) \leq\left(\tan \left(\frac{\pi T\left(\epsilon_{0}\right)}{2 t_{c}}\right)\right)^{1 / \kappa}
        \label{eq20}
    \end{equation}

This equation highlights that the term \((t /\) \(\left.t_{c}\right)^{\alpha-1} \geq 1\) in (\ref{eq14}) increases as \(\alpha\) decreases for any \(t \leq t_{c}\), suggesting that a smaller \(\alpha\) results in a shorter settling time \(T\left(\epsilon_{0}\right)\). Consequently, (\ref{eq20}) indicates that a smaller \(\alpha\) also leads to a smaller steady-state residual error \(\|\epsilon(\infty)\|\), thereby implying that the FOZNN model with a smaller \(\alpha\) achieves higher precision.
\end{remark}

\section{Experiments and Comparison}
\noindent
The effectiveness of the proposed PTC-NTFOZNN model is determined through comparative analysis with six established ZNN models \cite{11,18,19,20,38,39}, detailed in Table~\ref{tbl2}. Evaluations focus on resolving a TVQP example and include simulations and empirical testing related to cyclic motion planning tasks using the Flexiv Rizon robotic arm.

\subsection{An Illustrative Example}
\noindent
Typically, validating the proposed PTC-NTFOZNN model via a numerical test case enhances credibility. This section considers an illustrative example by specifying values for the coefficient matrix in (\ref{eq2}). The assignment of the coefficient matrix is as follows: \(A=[\sin (2 t), \cos (2 t)]\), \(\boldsymbol{b}=\sin (3 t)\), \(C=[I,-I]^{\mathrm{T}}\), \(\boldsymbol{d}=[1,1,1,1]^{\mathrm{T}}\), \(\boldsymbol{p}=\left[\cos(2t),\sin(2t)\right]^\mathrm{T}\), and
\[
\Omega=\left[\begin{array}{cc}
\cos (t) / 4+1 & \sin (t) / 4 \\
\sin (t) / 4 & \sin (t) / 4+1
\end{array}\right]
\]
where \(I\) represents the identity matrix.

Due to the lack of an explicit-form theoretical solution for the specified problem, the \textit{quadprog} function in Matlab is employed to derive a high accuracy numerical solution, achieving an error margin of \(10^{-15}\). This numerical solution is represented as \(\boldsymbol{y}^{*}(t)=\left[y_{1}^{*}(t), y_{2}^{*}(t)\right]\), and it acts as the reference standard for comparative analysis across seven neural models. For consistency in assessment, each neural model is configured with identical parameters: a gain factor \(\gamma=2\), a predefined time constant \(t_{c}=1\), \(r= 0.5\), \(\kappa=0.5\), and \(\xi=\zeta / \gamma\), where \(\zeta\) is set as five times the least upper bound of the additive noise. Furthermore, \(\varepsilon\) in the perturbed FB function is fixed to be \(1 \times 10^{-6}\).

\begin{figure}[t]
	\centering
		\includegraphics[scale=.32]{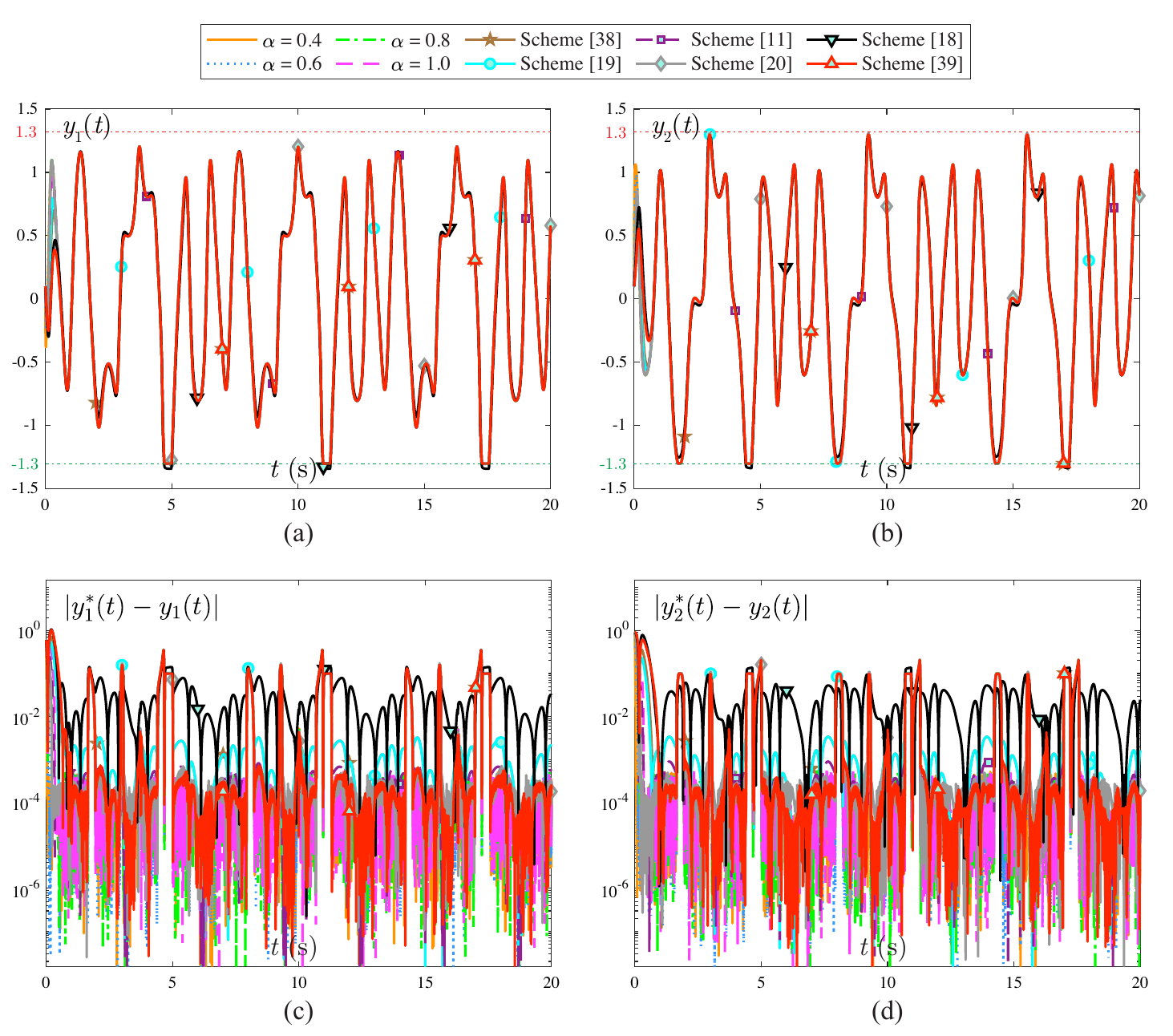}
	\caption{Neural states (a) \(y_{1}\), (b) \(y_{2}\), and their respective errors (c) \(\left|y_{1}^{*}-y_{1}\right|\), (d) \(\left|y_{2}^{*}-y_{2}\right|\) for six established models, compared to our PTC-NT-FOZNN at four \(\alpha\) values when the additive noise \(\boldsymbol{\delta}(t)=\) \(0.1 \sin (t)\) is considered, as tackling the case from Section 4.1.}
	\label{fig1}
\end{figure}

\begin{figure}[t]
	\centering
		\includegraphics[scale=.32]{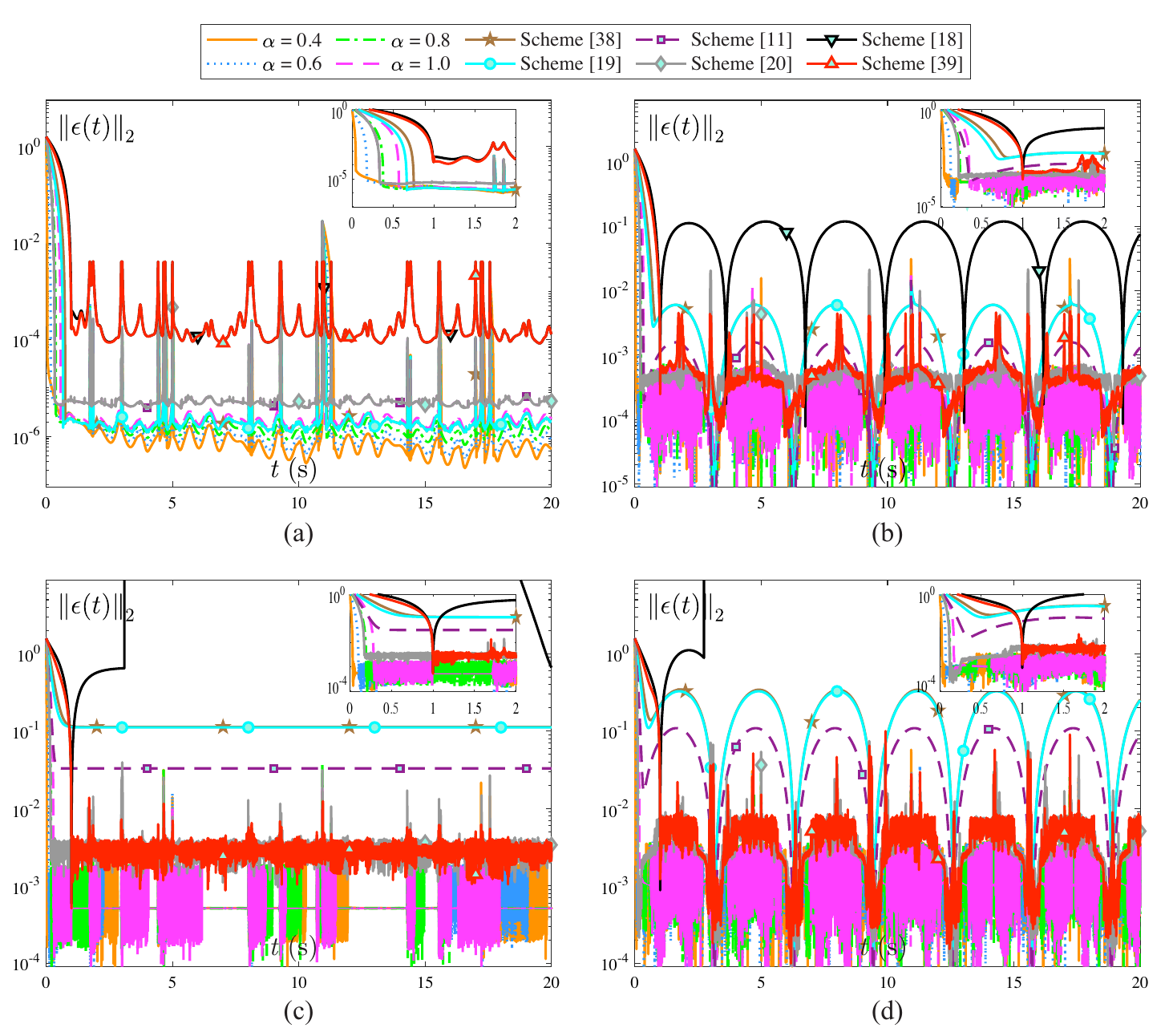}
	\caption{Residual error \(\|\boldsymbol{\epsilon}(t)\|_{2}\) for six established models compared to our PTC-NT-FOZNN at four \(\alpha\) values under various noise conditions: (a) no additive noise, (b) \(\boldsymbol{\delta}(t)=0.1 \sin (t)\), (c) constant noise \(\boldsymbol{\delta}(t)=\) 0.5, and (d) \(\boldsymbol{\delta}(t)=\sin (t)\), as tackling the case from Section 4.1.}
	\label{fig2}
\end{figure}

As detailed in Table~\ref{tbl2}, our PTC-NT-FOZNN model incorporates a time-dependent variable gain, \(\gamma(t)=\gamma t^{\alpha-1}\), where \(0<\alpha \leq 1\). This modification of the gain factor is designed to improve the standard ZNN model by rapidly reducing the gain, thereby lowering power consumption in hardware realization. The effectiveness of this approach in noise rejection was assessed using three types of additive noise: \(\boldsymbol{\delta}(t)=0.1 \sin (t)\), \(\boldsymbol{\delta}(t)=0.5\), and \(\boldsymbol{\delta}(t)=\sin (t)\). The resilience of the neural models to noise is demonstrated in Fig.~\ref{fig1}, which presents the neural state profiles and absolute state error trajectories when subjected to \(\boldsymbol{\delta}(t)=0.1 \sin (t)\). This figure highlights the robust noise handling capabilities of schemes \cite{20,39}, and our PTC-NT-FOZNN model, with positional errors maintained at a maximum of \(10^{-4}\).

Furthermore, Fig.~\ref{fig2} displays the residual error profiles for seven models, both under noise-free condition and the two specified noise environments. Remarkably, the PTC-NT-FOZNN model, particularly at four distinct values of \(\alpha\), exhibits exceptional precision and improved noise immunity relative to its counterparts. Notably, this model achieves convergence (i.e., \(\epsilon(t) \sim 10^{-4}\) ) within a predefined time of 1 second in scenarios with \(\boldsymbol{\delta}(t)=0.1 \sin (t)\), reducing the residual error by over tenfold compared to scheme \cite{39} and other neural models. These empirical results confirm the superior performance of the PTC-NT-FOZNN model in handling TVQP problems.

\subsection{Simulative Verification}
\noindent
This section outlines the utilization of the PTC-NT-FOZNN in cyclic motion planning for industrial robotic arms. The kinematics-level cyclic motion planning of a redundant robotic arm can be formulated as the following TVQP problem \cite{18,40}:
\begin{equation}
\begin{array}{ll}
\min & \dot{\boldsymbol{\theta}}^{\mathrm{T}}(t) \dot{\boldsymbol{\theta}}(t) / 2+\boldsymbol{\psi}^{\mathrm{T}}(t) \dot{\boldsymbol{q}}(t) \\
\text { s.t. } & J(t) \dot{\boldsymbol{\theta}}(t)=\dot{\boldsymbol{w}}(t) \\
& \boldsymbol{\theta}^{-} \leq \boldsymbol{\theta}(t) \leq \boldsymbol{\theta}^{+} \\
& \dot{\boldsymbol{\theta}}^{-} \leq \dot{\boldsymbol{\theta}}(t) \leq \dot{\boldsymbol{\theta}}^{+}
\end{array}
\label{eq21}
\end{equation}

In this formulation, \(\boldsymbol{\theta}(t) \in \mathbb{R}^{n}\) and \(\dot{\boldsymbol{\theta}}(t) \in \mathbb{R}^{n}\) denote the joint angles and velocities of the robotic arm, respectively. The associated vector \(\boldsymbol{\psi}(t)=\mu(\boldsymbol{\theta}(t)- \boldsymbol{\theta}_{0})\) facilitates the repetitive movement of the robotic arm, where \(\mu \in \mathbb{R}\) is a positive constant factor and \(\theta_{0}\) represents the initial joint angles. \(J(t) \in \mathbb{R}^{3 \times n}\) is defined as the Jacobian matrix, and \(\boldsymbol{\omega}(t) \in \mathbb{R}^{3}\) represents the desired trajectory of the end-effector. The limits \(\boldsymbol{\theta}^{-}\), \(\boldsymbol{\theta}^{+}\), \(\dot{\boldsymbol{\theta}}^{-}\)and \(\dot{\boldsymbol{\theta}}^{+}\) define the allowable ranges for the joint angles and velocities. The problem is then expressed in the condensed form (\ref{eq2}) where \(\boldsymbol{y}(t)=\dot{\boldsymbol{\theta}}(t)\), \(\Omega(t)=I\), \(A(t)=J(t)\), \(\boldsymbol{b}(t)=\dot{\boldsymbol{\omega}}(t)\), \(C(t)=[I,-I]^{\mathrm{T}}\), and \(\boldsymbol{d}(t)=[\boldsymbol{d}^{+\mathrm{T}},-\boldsymbol{d}^{-\mathrm{T}}]^\mathrm{T}\). Besides, according to \cite{14,18}, additional functions are incorporated to ensure the smoothness of \(\boldsymbol{d}(t)\), i.e.,
\[
\begin{aligned}
& \boldsymbol{d}^{-}= \begin{cases}\dot{\boldsymbol{\theta}}^{-}, & \text {if } \boldsymbol{\theta}(t) \in\left[\boldsymbol{\eta}_{\mathbf{1}}, \boldsymbol{\theta}^{+}\right] \\
\dot{\boldsymbol{\theta}}^{-}\left(1-f_{1}(\boldsymbol{\theta}(t))\right), & \text {if } \boldsymbol{\theta}(t) \in\left[\boldsymbol{\theta}^{-}, \boldsymbol{\eta}_{\mathbf{1}}\right]\end{cases} \\
& \boldsymbol{d}^{+}= \begin{cases}\dot{\boldsymbol{\theta}}^{+}, & \text {if } \boldsymbol{\theta}(t) \in\left[\boldsymbol{\theta}^{-}, \boldsymbol{\eta}_{\mathbf{2}}\right] \\
\dot{\boldsymbol{\theta}}^{+}\left(1-f_{2}(\boldsymbol{\theta}(t))\right), & \text {if } \boldsymbol{\theta}(t) \in\left[\boldsymbol{\eta}_{\mathbf{2}}, \boldsymbol{\theta}^{+}\right]\end{cases}
\end{aligned}
\]
where\\
\(f_{1}(\boldsymbol{\theta}(t))=(\sin (0.5 \pi(\sin (0.5 \pi(\boldsymbol{\theta}(t)-\boldsymbol{\eta}_{\mathbf{1}}) / \boldsymbol{\eta}_{\mathbf{3}}))^{2}))^{2}\), \(f_{2}(\boldsymbol{\theta}(t))=(\sin (0.5 \pi(\sin (0.5 \pi(\boldsymbol{\theta}(t)-\boldsymbol{\eta}_{\mathbf{2}}) / \boldsymbol{\eta}_{\mathbf{4}}))^{2}))^{2}\), \(\boldsymbol{\eta}_{\boldsymbol{1}}=\kappa_{1} \boldsymbol{\theta}^{-}\), \(\boldsymbol{\eta}_{\boldsymbol{2}}=\kappa_{2} \boldsymbol{\theta}^{+}\), \(\boldsymbol{\eta}_{\mathbf{3}}=\boldsymbol{\theta}^{-}-\boldsymbol{\eta}_{\boldsymbol{1}}\), and \(\boldsymbol{\eta}_{\boldsymbol{4}}= \boldsymbol{\theta}^{+}-\boldsymbol{\eta}_{\boldsymbol{2}}\), with \(0<\kappa_{1}, \kappa_{2}<1\) being two positive constants.

\begin{figure}[t]
	\centering
		\includegraphics[scale=.32]{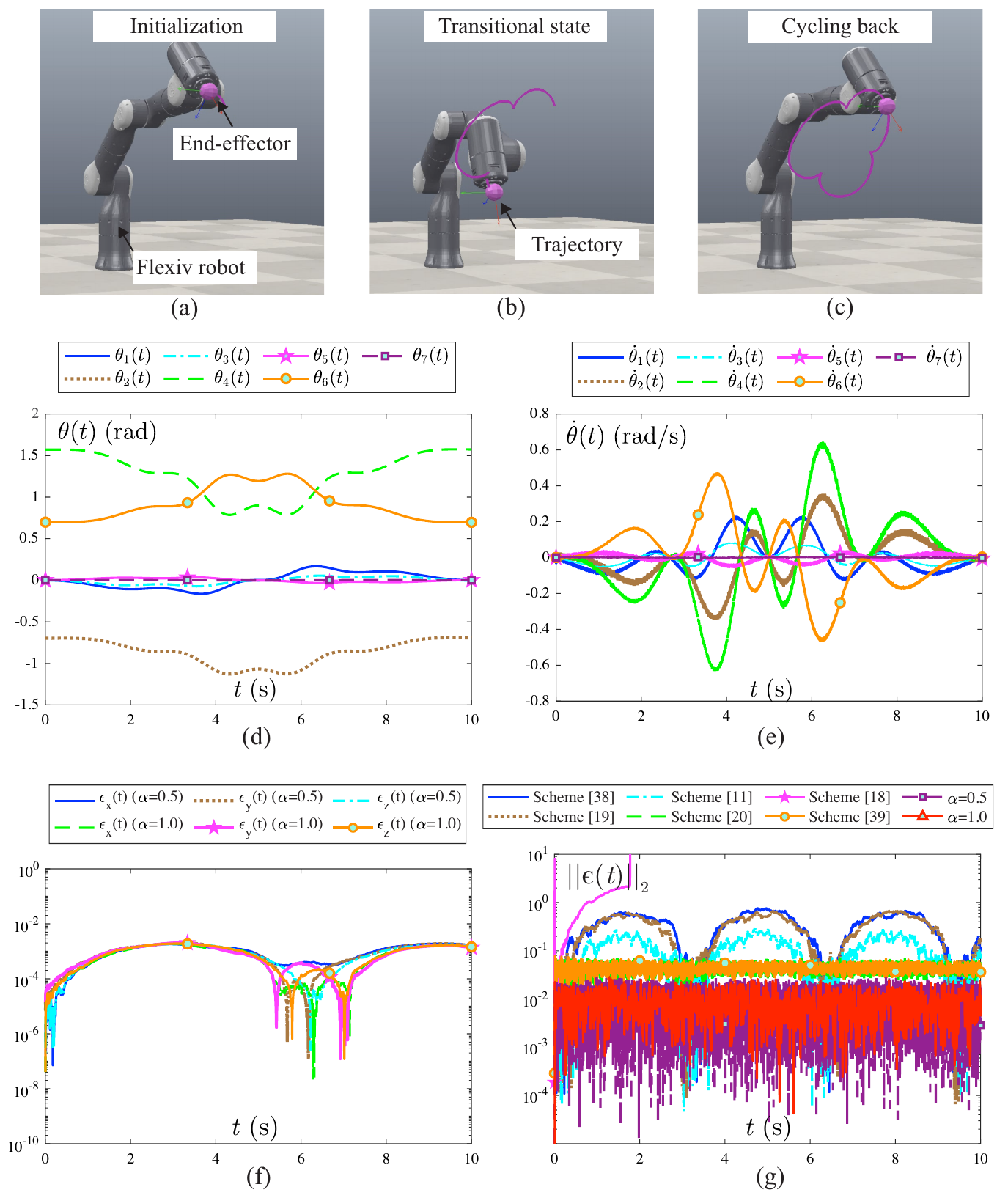}
	\caption{Illustration of the captured phases: (a) initialization, (b) transitional state, and (c) cycling-back state for a simulated Flexiv robot tracking a clover-shaped path, and the simulated (d) joint angles, (e) joint velocities, (f) absolute tracking errors calculated using the PTC-NT-FOZNN model, and (g) residual errors evaluated by seven different neural models under an uncertain noise \(\boldsymbol{\delta}(t)=\sin (t)+\cos (t)+\bar{n}(t)\).}
	\label{fig3}
\end{figure}

\begin{figure}[t]
	\centering
		\includegraphics[scale=.32]{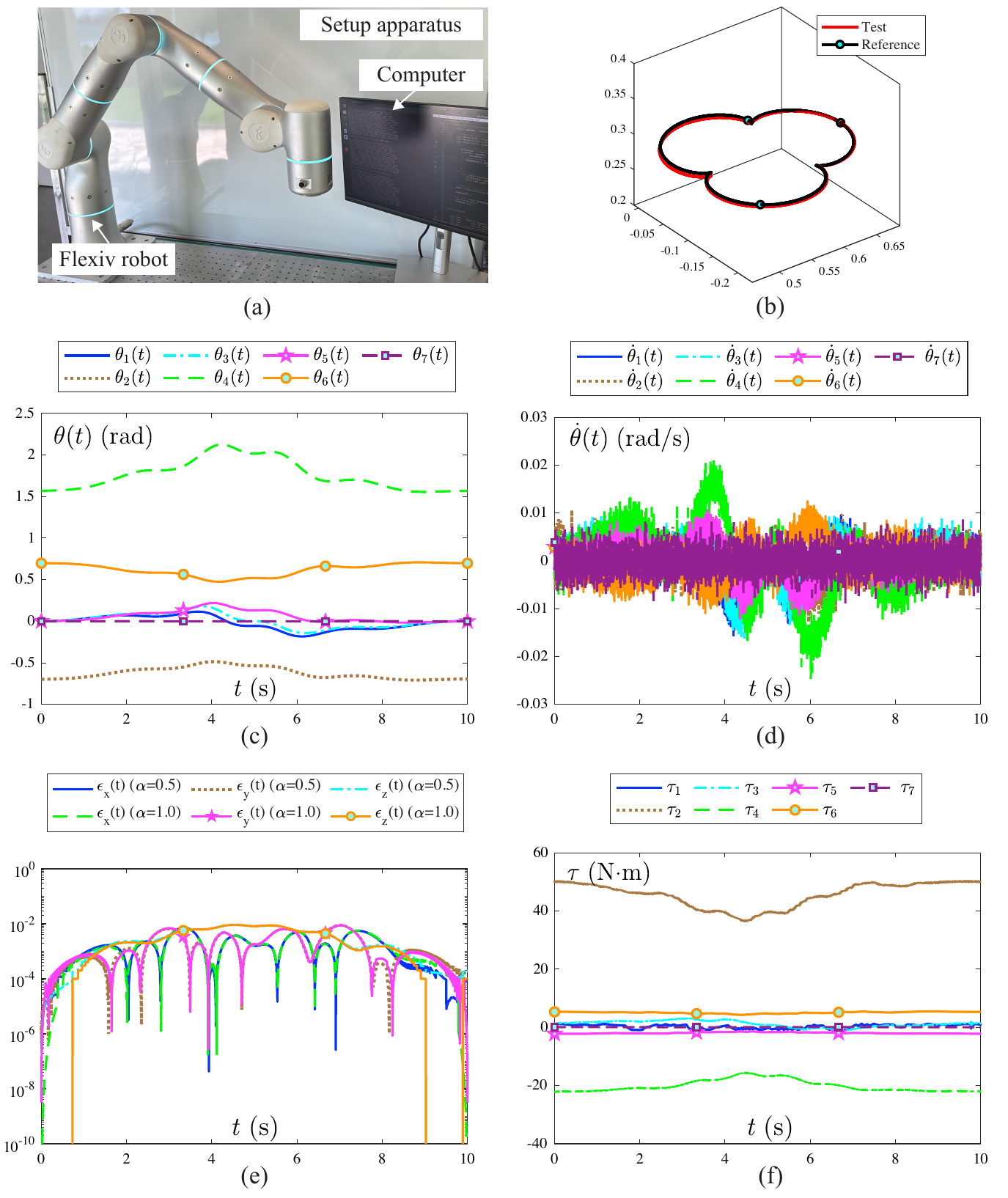}
	\caption{Illustration of (a) an image of the experimental setup, (b) a comparison between the actual trajectory and the intended clovershaped path, along with (c) joint angles, (d) joint velocities, (e) absolute tracking errors, and (f) real-time joint torques as measured during experiments using the PTC-NT-FOZNN model on the physical test platform.}
	\label{fig4}
\end{figure}

In this simulation, the Flexiv Rizon robotic arm, equipped with seven joints, is deployed using the CoppeliaSim software \cite{41} for executing a cyclic motion planning task. The control mechanism is governed by the PTC-NT-FOZNN model, which is specifically set with \(\alpha=0.5\) and \(\alpha=1.0\), and targets a fast convergence within a predefined 0.1 seconds. This configuration effectively addresses the computational uncertainties caused by an additive noise \(\boldsymbol{\delta}(t)=\sin (t)+\cos (t)+\bar{n}(t)\), with \(\bar{n}(t)\) being a white noise signal bounded by $2$. The primary objective for the robotic arm is to precisely navigate a clover-shaped trajectory.

Fig.~\ref{fig3}(a)-(c) visually captures the results of the simulation, demonstrating the end-effector's adeptness at accurately following the predetermined path. Fig.~\ref{fig3}(d)-(e) further elucidate the joint angles and velocities achieved by the PTC-NT-FOZNN model with \(\alpha= 0.5\), illustrating a successful reset to initial positions, thereby affirming the model's efficiency in precise cyclic motion planning. This model maintains all joint movements strictly within the pre-specified limits, ensuring tracking errors remain below \(10^{-3}\) meters. Additionally, Fig.~\ref{fig3}(g) displays the residual errors \(\|\boldsymbol{\epsilon}(t)\|_{2}\) among seven different neural network models, confirming the PTC-NT-FOZNN model's conformity to the predefined convergence timelines as postulated in Section 3.2. The comparative analysis accentuates its exceptional tracking accuracy and resilience to additive noise, proving its efficacy for complex robotic tasks.

\subsection{Practical Verification}
\noindent
The PTC-NT-FOZNN model's utility is further manifested through rigorous experimental testing with a Flexiv Rizon robot, programmed to execute cyclical movements following a clover-shaped path as illustrated in Fig.~\ref{fig4}(a). The demonstration video is available from the URL:\\
\text{https://drive.google.com/file/d/12IDBScPH\_9z-SF6GoL}\\
\text{-g\_7MT8yB8OJa6/view?usp=sharing}

In the context of motion planning, the Flexiv robot's end-effector position is specified with precision, while the other degrees of freedom remain unconstrained. The range of joint angles for the 7-joint Flexiv Rizon robot arm is set as follows:\\
Lower limit \(\left(\theta^{-}\right)\):\\
\(-\left[161, 131.5, 172.5, 107, 172.5, 82.5, 172.5\right]^{\mathrm{T}}\) deg\\
Upper limit \(\left(\theta^{+}\right)\):\\
\(+\left[161, 131.5, 172.5, 155, 172.5, 262.5, 172.5\right]^{\mathrm{T}}\) deg

These limits are consistent with the specified joint angle range for the real Flexiv Rizon robot arm. Additionally, the joint-velocity bounds are established at \(\dot{\theta}^{-}=-0.65\) rad/s and \(\dot{\theta}^{+}=0.65\) rad/s, with a scaling factor \(\mu\) set to $1$. This setup ensures that the robot operates within safe and efficient kinematic parameters.

The joint angles and velocities recorded, shown in Fig.~\ref{fig4}(c)-(d), verify the robot's ability to cyclically revert to its original positions. Fig.~\ref{fig4}(b) displays the three-dimensional trajectory of the robot's end-effector, compared with the designated path, emphasizing path-tracking precision within a range of \(10^{-4}\) to \(10^{-2}\) meters, consistent with the positional accuracy depicted in Fig.~\ref{fig4}(e). Additionally, Fig.~\ref{fig4}(f) details the joint torques observed during these trials, reflecting the robot's stable and smooth motion, key for accurate path tracking. These findings collectively support the PTC-NT-FOZNN model's effectiveness and suitability for robotic motion planning tasks.

\section{Conclusion}
\noindent
A PTC-NT-FOZNN model, specifically designed for addressing TVQP challenges, is proposed, and studied in this paper with a focus on its utility in cyclic motion planning for robotic systems. Through theoretical analysis, it is established that the PTC-NTFOZNN model not only converges within a predefined time but also demonstrates robustness to noise. A comparative analysis, utilizing an illustrative TVQP example, shows that the PTC-NT-FOZNN model surpasses six other representative ZNN models in both convergence accuracy and robustness. Furthermore, the model's practicality is validated through its successful deployment in the cyclic motion planning of a Flexiv robotic arm, highlighting its effectiveness as a robotic inverse kinematic solver. This development points towards exciting possibilities for future ZNN hardware solutions that are both energy-efficient and optimized for predefined-time convergence.

\vskip 2mm
\large
\noindent
\textbf{Acknowledgment}
\vskip 2mm

\Acknow
\noindent
This work was supported in part by the National Natural Science Foundation of China under Grant 52205032, in part by the Shun Hing Institute of Advanced Engineering, The Chinese University of Hong Kong, in part by Research Grants Council of Hong Kong (Ref. No. 14204423), and in part by Guangdong Basic and Applied Basic Research Foundation under Grant 2023A1515010062.

\vskip 2mm
\renewcommand\refname{\large

\end{document}